\documentclass[conference]{IEEEtran}
\IEEEoverridecommandlockouts
\usepackage{cite}
\usepackage{amsmath,amssymb,amsfonts}
\usepackage[nolist]{acronym}
\usepackage{algorithmic}
\usepackage{booktabs}
\usepackage[caption=false,font=footnotesize]{subfig}
\usepackage{graphicx}
\usepackage{textcomp}
\usepackage{xcolor}
\usepackage{orcidlink}
\usepackage{custom_commands}
\usepackage{mwe}
\usepackage[capitalize]{cleveref}
\usepackage{siunitx}
\usepackage{tikz}

\usetikzlibrary{calc,arrows.meta,angles,quotes,positioning}

\hypersetup{
    colorlinks = true,
    citecolor = {black},
    linkcolor = {black},
    urlcolor = {black},
}

\makeatletter
\let\ACplacelabel\AC@placelabel
\makeatother

\def\BibTeX{{\rm B\kern-.05em{\sc i\kern-.025em b}\kern-.08em
    T\kern-.1667em\lower.7ex\hbox{E}\kern-.125emX}}
\begin{document}

\title{Context-Aware Sensor Modeling for Asynchronous Multi-Sensor Tracking in Stone Soup\\
\thanks{This work was funded in part by the Situational Awareness programme (P1756) at the Norwegian Defence Research Establishment (FFI), and by Universitetssenteret på Kjeller (UNIK).}
}

\newcommand{\sharedaffiliation}{
Norwegian Defence Research Establishment (FFI)\\
University of Oslo, Dept. of Technology Systems
}

\author{
\IEEEauthorblockN{
Martin Vonheim Larsen\IEEEauthorrefmark{1}\IEEEauthorrefmark{2} \orcidlink{0000-0002-3008-7712},
Kim Mathiassen\IEEEauthorrefmark{1}\IEEEauthorrefmark{2} \orcidlink{0000-0003-3747-5934}
}
\IEEEauthorblockA{\IEEEauthorrefmark{1}
Norwegian Defence Research Establishment (FFI)}
\IEEEauthorblockA{\IEEEauthorrefmark{2}
University of Oslo, Department of Technology Systems}
}

\maketitle

\begin{abstract}
Multi-sensor tracking in the real world involves asynchronous sensors with partial coverage and heterogeneous detection performance.
Although probabilistic tracking methods permit detection probability and clutter intensity to depend on state and sensing context, many practical frameworks enforce globally uniform observability assumptions.
Under multi-rate and partially overlapping sensing, this simplification causes repeated non-detections from high-rate sensors to erode tracks visible only to low-rate sensors, potentially degrading fusion performance.

We introduce \verb#DetectorContext#, an abstraction for the open-source multi-target tracking framework Stone Soup.
\verb#DetectorContext# exposes detection probability and clutter intensity as state-dependent functions evaluated during hypothesis formation.
The abstraction integrates with existing probabilistic trackers without modifying their update equations.
Experiments on asynchronous radar-lidar data demonstrate that context-aware modeling restores stable fusion and significantly improves HOTA and GOSPA performance without increasing false tracks.
\end{abstract}


\begin{acronym}
\acro{ANEES}{average normalized estimation error squared}
\acro{BA}{bundle adjustment}
\acro{DAQ}{dual absolute quadric}
\acro{DIAC}{dual image of the absolute conic}
\acro{DLT}{direct linear transform}
\acro{DoF}{degrees of freedom}
\acro{ESM}{electronic support measures}
\acroplural{ESM}{electronic support measures}
\acro{FFI}{Norwegian Defence Research Establishment}
\acro{HFOV}{horizontal field-of-view}
\acro{FOV}{field-of-view}
\acro{IMU}{inertial measurement unit}
\acro{LM}{Levenberg-Marquardt}
\acro{PTZ}{pan-tilt-zoom}
\acro{MAP}{maximum a posteriori}
\acro{MAE}{mean absolute error}
\acro{MRE}{mean relative error}
\acro{MC}{Monte Carlo}
\acro{MEPE}{mean estimated projection error}
\acro{OE}{orientation estimation}
\acro{PT}{pan/tilt}
\acro{PTCEE}{pan/tilt camera extrinsic and intrinsic estimation}
\acro{FRD}{forward-right-down}
\acro{RDF}{right-down-forward}
\acro{RMSE}{root mean square error}
\acro{RS}{rolling shutter}
\acro{SLAM}{simultaneous localization and mapping}
\acro{UiO}{University of Oslo}
\acro{VO}{visual odometry}
\acro{VIO}{visual-inertial odometry}
\acro{GNSS}{Global Navigation Satellite System}
\acro{MCM}{mine countermeasures}
\acro{SA}{situational awareness}
\acro{ASV}{autonomous surface vessel}
\acro{USV}{unmanned surface vessel}
\acro{POTT}{proportion-of-time-tracked}
\acro{TTT}{time-to-track}
\acro{IDSW}{number of ID switches}
\acro{RTK}{real-time kinematics}
\acro{INS}{Inertial Navigation System}
\acro{RTK}{Real-Time Kinematic}
\acro{JPDA}{Joint Probabilistic Data Association}
\acro{GM-PHD}{Gaussian-mixture Probability Hypothesis Density}
\acro{HOTA}{Higher Order Tracking Accuracy}
\acro{GOSPA}{Generalized Optimal Sub-pattern Assignment}
\acro{PMBM}{Poisson Multi-Bernoulli Mixture}
\acro{LMB}{Labeled Multi-Bernoulli}
\acro{RFS}{Random Finite Set}
\acro{OOSM}{out-of-sequence modeling}
\acro{STO}{Science and Technology Organization}
\end{acronym}
\section{Introduction}\label{sec:intro}
Real-world multi-sensor tracking operates on data streams that are asynchronous, heterogeneous, and partially overlapping~\cite{bar-shalom_multitarget-multisensor_1995}.
In such environments, the observability of targets varies across the surveillance volume: both detection probability and clutter intensity depend on sensing geometry, scene structure, and sensor characteristics~\cite{wilthil_estimation_2019}.
Although Bayesian tracking theory allows detection probability ($P_D$) and clutter intensity ($\lambda$) to depend on state and context~\cite{bar-shalom_multitarget-multisensor_1995,mahler_statistical_2007}, many practical tracking frameworks reduce them to global constants.
As a result, implementations assume uniform observability across the surveillance volume.
This simplification prevents the tracker from representing where detection is likely and where clutter is expected.

Globally uniform observability assumptions fail under partially overlapping and asynchronous sensing.
Consider a long-range radar updating at 0.5 Hz fused with a short-range lidar updating at 10 Hz.
Between two radar detections, the fused tracker processes 20 lidar updates.
Once a target exits lidar coverage, these updates appear as consecutive missed detections.
In probabilistic trackers with a globally constant observability model, each missed detection contributes a factor proportional to $(1 - P_D)$ in the update step~\cite{bar-shalom_multitarget-multisensor_1995}.
As a result, the track existence probability decays multiplicatively across updates.
To prevent premature deletion, the tracker must therefore be tuned to tolerate sustained sequences of missed detections.
This relaxation in turn allows the survival of short-range clutter tracks, which a dedicated lidar tracker would trivially reject.
The fused system consequently performs worse than its parts.

This failure mode is particularly acute in rigorous probabilistic trackers.
These algorithms model track existence continuously~\cite{bar-shalom_probabilistic_2009,mahler_statistical_2007}, allowing biases in detection and clutter modeling to accumulate.
Heuristic trackers with hard confirmation logic can be less sensitive to persistent modeling errors.
However, probabilistic methods represent observability explicitly.
With an appropriate sensor model, this failure mode can therefore be avoided.

Although state-dependent sensor models appear in specialized implementations, general-purpose tracking frameworks do not expose this flexibility.
Stone Soup is an open-source, general-purpose multi-target tracking framework~\cite{hiscocks_stone_2023}.
In the current Stone Soup interfaces, detection probability and clutter intensity are defined as constant scalar parameters on the hypothesizer or updater (e.g., in \verb#PDAHypothesiser#, \verb#PHDUpdater# or \verb#BernoulliParticleUpdater#).
Detection probability is broadcast uniformly across predicted states, and clutter intensity appears as a fixed normalization term in hypothesis weighting.
This design enforces globally uniform observability within existing updater logic; modeling state-dependent $P_D(x)$ or $\lambda(z)$ requires refactoring core update code rather than supplying richer sensor models.

This limitation reflects a missing abstraction boundary between sensor modeling and hypothesis evaluation.
Although probabilistic trackers already incorporate detection probability and clutter intensity in their update equations, the framework does not expose these quantities as state- or measurement-dependent functions.
The required change is therefore architectural rather than mathematical.

We introduce \verb#DetectorContext#, a lightweight interface that provides per-hypothesis access to detection probability and clutter intensity during update.
By evaluating observability during hypothesis scoring rather than as global scalars, the abstraction restores state-dependent modeling without altering core tracker equations.
In practice, this allows detection probability and clutter intensity to depend on factors such as sensor range, blind sectors, or scene-dependent detector behavior, while leaving the underlying tracker equations unchanged.

This paper makes the following contributions:
\begin{itemize}
    \item \textbf{Characterization of global observability failure mode:}
    We demonstrate how globally uniform detection and clutter assumptions induce structural failure under asynchronous and partially overlapping sensing, and explain why this effect accumulates in probabilistic trackers.
    \item \textbf{Framework-level abstraction for context-aware observability:}
    We introduce \verb#DetectorContext#, an architectural interface for Stone Soup that exposes state- and measurement-dependent detection probability and clutter intensity during hypothesis evaluation without modifying tracker equations.
    \item \textbf{Empirical validation on asynchronous radar-lidar fusion:}
    We show that context-aware modeling restores effective fusion performance, improving track establishment while maintaining clutter suppression on a real-world maritime dataset.
\end{itemize}
\section{Target Observability Modeling in Multi-Target Tracking Frameworks}\label{sec:related-work}
This section briefly reviews how sensor observability is treated in probabilistic tracking theory and contrasts this with the abstractions exposed by general-purpose tracking frameworks.

\subsection{State-Dependent Observability in Tracking Theory}
Probabilistic multi-target tracking formulations include detection probability and clutter intensity as fundamental components of the measurement model.
In foundational treatments~\cite{bar-shalom_multitarget-multisensor_1995}, these quantities are typically presented as unspecified parameters.
While this formulation theoretically permits dependence on target state or sensing geometry, it does not prescribe a specific functional form.
In \ac{RFS} formulations, detection probability and clutter intensity are explicitly defined as state- and measurement-dependent functions, $P_D(\vecx)$ and $\lambda(\vecz)$~\cite{mahler_statistical_2007}. 
This structure is retained in modern \ac{RFS}-based filters such as the \ac{LMB} filter~\cite{reuter_labeled_2014} and the \ac{PMBM} filter~\cite{williams_marginal_2015}, where detection probability enters directly into hypothesis weights and existence updates.

Explicit modeling of state-dependent observability is rare in general-purpose tracking frameworks, though bespoke systems demonstrate its value.
Work in the mid-2000s recognized that constant detection performance is unrealistic in practice~\cite{bar-shalom_probabilistic_2009}.
Structured solutions have since emerged in maritime surveillance, where range-dependent clutter and occlusion demand explicit observability modeling~\cite{wilthil_estimation_2019,helgesen_heterogeneous_2022,brekke_multitarget_2021}.
These systems demonstrate that probabilistic trackers can exploit sensor context when available.
However, this modeling remains tightly coupled to specific implementations rather than exposed through reusable abstractions.

Handling asynchronous and multi-rate sensing has also been studied extensively, including out-of-sequence measurement (OOSM) processing and multi-rate update strategies~\cite{bar-shalom_estimation_2001}. 
These works address temporal misalignment and delayed measurements, but typically assume globally specified detection characteristics. 
The interaction between asynchronous sensing and spatially varying observability has received comparatively little attention at the framework abstraction level.

\subsection{Limitations of General-Purpose Tracking Frameworks}
Despite the theoretical flexibility of probabilistic tracking formulations, this flexibility is not typically exposed at the framework interface level.
In contrast to bespoke systems, general-purpose tracking frameworks prioritize modularity and algorithm reuse.
Within Stone Soup, detection probability and clutter intensity are specified as global quantities.
Currently, conditioning on predicted track state, measurement location, or instantaneous sensor coverage is not supported~\cite{hiscocks_stone_2023}.

This restriction is particularly limiting for probabilistic tracking algorithms, which consume detection probability and clutter intensity directly during hypothesis evaluation and track management.
When observability varies across the surveillance volume, a globally uniform model forces trackers to interpret all missed detections equivalently, regardless of whether the target was actually observable.
Under partially overlapping and asynchronous sensing, uniform treatment of missed detections induces the systematic failure modes described in \cref{sec:intro}.

Importantly, this limitation does not arise from the tracking algorithms themselves, but from the absence of a framework-level mechanism for exposing sensor context during hypothesis evaluation.
Addressing this gap requires an architectural solution that preserves algorithm modularity while enabling observability to be modeled locally and dynamically.
\section{The DetectorContext Abstraction}
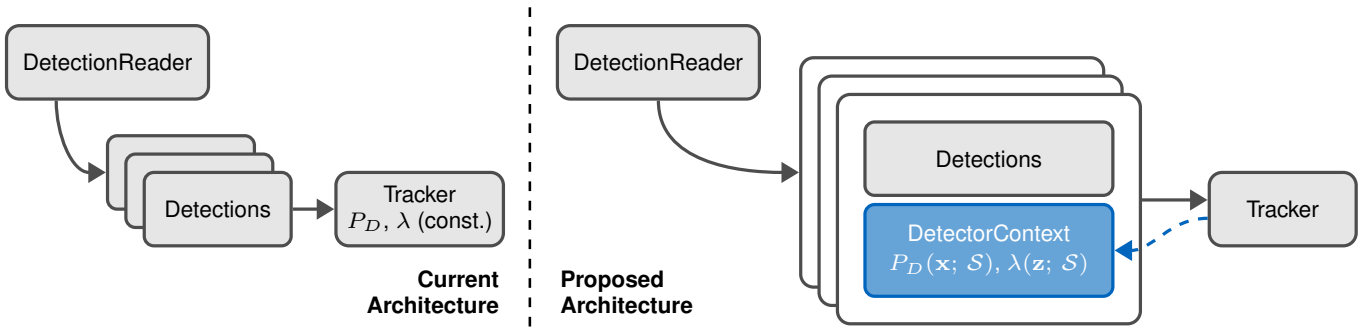
\begin{figure*}[t]
\centering
\resizebox{\textwidth}{!}{%
    \begin{tikzpicture}[
    font=\sffamily\scriptsize,
    box/.style={draw, rounded corners, minimum height=0.8cm, minimum width=1.6cm, align=center},
    stackbox/.style={draw, fill=white, rounded corners=3pt, minimum height=0.8cm, minimum width=1.6cm, align=center},
    bigbox/.style={draw, fill=white, rounded corners=4pt, minimum height=1.9cm, align=center},
    every path/.style={line width=1pt},
    >=Triangle,
    arrow/.style={->, black!70}
]
\definecolor{ffiWater}{RGB}{0,101,189}

\begin{scope}[xshift=-6.7cm]
\draw[dashed, thick] (5.2, 2.2) -- (5.2,-1.3);

\node[box,minimum width=2.2cm, fill=black!10, draw=black!70] (reader_old) at (0.6,1.6) {DetectionReader};

\node[stackbox, fill=black!10, draw=black!70] (det1_old) at (1.4,0.4) {};
\node[stackbox, fill=black!10, draw=black!70] (det2_old) at (1.6,0.2) {};
\node[stackbox, fill=black!10, draw=black!70] (det3_old) at (1.8,0.0) {Detections};

\draw[arrow] ($(reader_old.south west)!0.5!(reader_old.south)$) 
    to[out=-90,in=180] (det1_old.west);

\node[box, minimum width=1.8cm, fill=black!10, draw=black!70] (tracker_old) at (4.0,0.0)
{
Tracker\\
$P_D$, $\lambda$ (const.)%
};

\draw[arrow] (det3_old.east) -- (tracker_old.west);

\node[anchor=east, align=right] at (5.0,-0.9) {\textbf{Current}\\\textbf{Architecture}};
\node[anchor=west, align=left] at (5.4,-0.9) {\textbf{Proposed}\\\textbf{Architecture}};
\end{scope}

\begin{scope}[xshift=1.7cm]

\node[box, minimum width=2.2cm, fill=black!10, draw=black!70] (reader_old) at (-1.8,1.6) {DetectionReader};

\node[bigbox, minimum width=3.3cm, minimum height=2.5cm, fill=white, draw=black!70] (out1) at (1.4,0.4) {};
\node[bigbox, minimum width=3.3cm, minimum height=2.5cm, fill=white, draw=black!70] (out2) at (1.6,0.2) {};
\node[bigbox, minimum width=3.3cm, minimum height=2.5cm, fill=white, draw=black!70] (out3) at (1.8,0.0) {};
\node[stackbox, minimum width=2.7cm, draw=black!70, fill=black!10] () at (1.8,0.55) {Detections};
\node[stackbox, minimum width=2.7cm, minimum height=1.0cm, fill=ffiWater!60, draw=ffiWater, text=white] (detctx) at (1.8,-0.45) {%
DetectorContext\\[1pt]
$P_D(\vecx;\, \mathcal{S})$, $\lambda(\vecz;\, \mathcal{S})$%
};

\draw[arrow] (reader_old.south) 
    to[out=-90,in=180] (out1.west);

\node[box, fill=black!10, draw=black!70] (tracker_old) at (5.0,0.0)
{Tracker};

\draw[arrow] ($(out3.east)+(0,0.1)$) -- ($(tracker_old.west)+(0,0.1)$);
\draw[arrow, dashed, ffiWater] ($(tracker_old.west)+(0,-0.1)$) to[out=180,in=0] (detctx.east);
\end{scope}

\end{tikzpicture}%
}
\caption{Architectural comparison. %
         \textbf{Left:} The tracker is configured with global detection probability $P_D$ and clutter intensity $\lambda$. %
         \textbf{Right:} The detection reader emits a per-timestep DetectorContext that exposes state and scene dependent functionals $P_D(\vecx;\ \mathcal{S})$ and $\lambda(\vecz;\ \mathcal{S})$ queried by the tracker.
         The key difference is that observability is no longer configured as static tracker parameters but evaluated dynamically via the emitted DetectorContext.}
\label{fig:architecture-diagram}
\end{figure*}

To address the architectural limitation identified in \cref{sec:related-work}, we introduce \verb#DetectorContext#, a framework-level abstraction for context-aware sensor modeling in Stone Soup.
This abstraction replaces globally specified detection probability and clutter intensity with locally conditioned evaluations during hypothesis formation.
The structural distinction between the current and proposed architectures is illustrated in \cref{fig:architecture-diagram}.

\subsection{The DetectorContext Interface}
The \verb#DetectorContext# interface encapsulates the sensor's instantaneous observability state.
At each timestep, the abstraction exposes two callable functions:
\begin{itemize}
    \item $P_D(\vecx;\,\mathcal{S})$: Returns the detection probability for a predicted track state $\vecx$.
    \item $\lambda(\vecz;\,\mathcal{S})$: Returns the clutter intensity at a measurement state $\vecz$.
\end{itemize}
Here, $\mathcal{S}$ represents the sensor state for the current timestep, such as platform pose, scanning geometry, coverage maps, blind sectors, or scene-dependent detector characteristics.

During hypothesis evaluation, trackers use $P_D(\vecx;\,\mathcal{S})$ and $\lambda(\vecz;\,\mathcal{S})$ in both detection and missed-detection hypotheses.
The interface aligns directly with standard probabilistic update equations and replaces globally specified constants with locally conditioned terms.

\subsection{Integration with Existing Trackers}
Most probabilistic tracking algorithms in Stone Soup already incorporate detection probability and clutter intensity in likelihood computation and existence updates.
By replacing globally specified constants with \verb#DetectorContext# queries, these algorithms can operate under locally conditioned observability without structural modification.

The abstraction is independent of the specific tracker formulation and preserves modular separation between sensor modeling and track management.
The \verb#DetectorContext# implementation is being prepared for upstream integration into the Stone Soup project repository and will be made publicly available.

\section{Experimental Evaluation}
We evaluate \verb#DetectorContext# on a real asynchronous radar-lidar dataset collected from an \ac{USV}.
The objective is to quantify the impact of globally uniform observability assumptions and to assess whether context-aware modeling restores effective sensor fusion.
All experiments are conducted within Stone Soup. 

\subsection{Dataset and Sensor Configuration}
The dataset was recorded using an Odin-class \ac{USV} equipped with a Simrad HALO-3 X-band pulse-compression radar and an Ouster OS2-128 lidar~\cite{larsen_warpath_2024}.
A navigation-grade \acs{INS} (\textit{NavP}, developed by FFI) and per-vessel \acs{RTK}-\acs{GNSS} loggers provide ground-truth trajectories for the ego vessel and all moving targets.

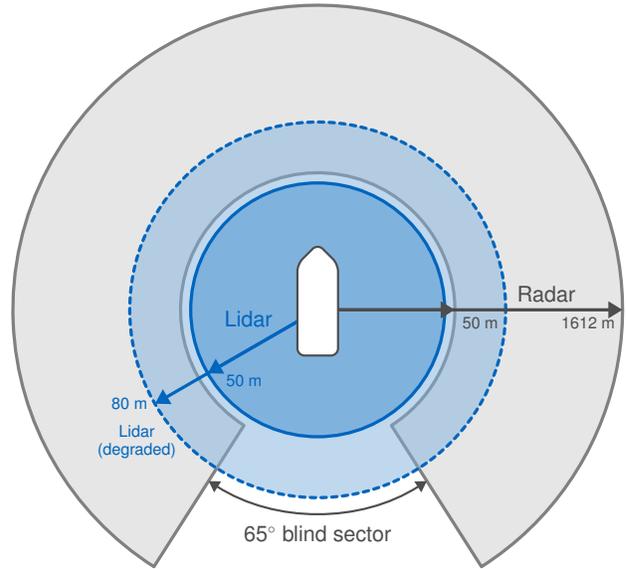
\begin{figure}[t]
    \centering
    \resizebox{\columnwidth}{!}{%
        \begin{tikzpicture}[
  scale=1,
  line cap=round,
  line join=round,
  font=\sffamily\small,
  every node/.style={font=\sffamily\small},
  >=Triangle,
  every path/.style={line width=1.8pt}
]
  \definecolor{ffiBerries}{RGB}{196,32,50}
  \definecolor{ffiWater}{RGB}{0,101,189}
  
  % ----------------------------
  % "Not to scale" radii (in cm)
  % ----------------------------
  \def\rLidarCrisp{2.5}   % represents 50 m
  \def\rLidarFaint{3.7}   % represents 80 m
  \def\rRadarInner{2.7}   % represents 50 m (same as lidar crisp)
  \def\rRadarOuter{6.0}   % represents 1612 m (NOT to scale)
  \def\blindHalf{32.5}    % half of 65 deg
  \def\blindCenter{270}   % rearward direction (boat points to +y)

  % Derived angles for blind sector
  \pgfmathsetmacro{\blindStart}{\blindCenter-\blindHalf} % 147.5
  \pgfmathsetmacro{\blindEnd}{\blindCenter+\blindHalf}   % 212.5

  % ----------------------------
  % coverage
  % ----------------------------

  \path[fill=black!10, draw=black!50]
    (\blindEnd:\rRadarOuter)
      arc[start angle=\blindEnd, end angle=\blindStart+360, radius=\rRadarOuter]
    -- (\blindStart+360:\rRadarInner)
      arc[start angle=\blindStart+360, end angle=\blindEnd, radius=\rRadarInner]
    -- cycle;
    
  \fill[ffiWater!50, opacity=0.5] (0,0) circle (\rLidarFaint);
  \draw[ffiWater, dashed] (0,0) circle (\rLidarFaint);
  
  \fill[ffiWater!50, opacity=1] (0,0) circle (\rLidarCrisp);
  \draw[ffiWater] (0,0) circle (\rLidarCrisp);

  % ----------------------------
  % Boat (3 x 9 rectangle + triangle on top), centered at origin
  % ----------------------------
  \def\boatW{0.4}  % half-width in cm  (so total 1.8)
  \def\boatH{0.9}  % half-height in cm (so total 5.4)  ~ "tall"
  \def\noseH{0.4}  % triangle height

  % ----------------------------
  % Annotations: ranges + blind sector angle
  % ----------------------------
  % Lidar range annotations (radial arrows)
  \draw[->, ffiWater] (0,0) -- (210:\rLidarFaint)
    node[at end, below left, align=center, xshift=16pt, yshift=-8pt] {\small{Lidar}\\\small(degraded)}
    node[at end, left] {\small80 m};
  \draw[->, ffiWater] (0,0) -- (210:\rLidarCrisp)
    node[pos=0.5, above, xshift=-8pt, yshift=4pt] {\large Lidar}
    node[pos=0.9, below right] {\small 50 m};

  % Radar range annotations (inner/outer)
  \draw[->, black!70] (0,0) -- (0:\rRadarInner)
    node[at end, below right] {\small 50 m};
  \draw[->, black!70] (0,0) -- (0:\rRadarOuter)
    node[pos=0.75, above] {\large Radar}
    node[at end, below left] {\small 1612 m};

  % Blind sector annotation (arc + label)
  \draw[very thick, black!70, <->]
    (\blindStart:4) arc[start angle=\blindStart, end angle=\blindEnd, radius=4];
  \node[black!70] at (\blindCenter:4.4) {\large 65$^\circ$ blind sector};
  
  % Hull
  \filldraw[white, draw=black!70, line width=1.2pt, rounded corners=4pt]
  (-\boatW,-\boatH) --
  (\boatW,-\boatH) --
  (\boatW,\boatH) --
  (0,\boatH+\noseH) --
  (-\boatW,\boatH) --
  cycle;

\end{tikzpicture}%
    }
    \caption{Effective sensing regions and blind sectors for the radar-lidar configuration.}
    \label{fig:sensor-coverage}
\end{figure}

The radar updates at 0.8~Hz and covers 50--1612~m, with a $65\degree$ rear blind sector.
The lidar updates at 10~Hz and provides reliable detections up to 50~m, with degraded coverage out to 80~m.
As illustrated in \cref{fig:sensor-coverage}, the sensors exhibit partially overlapping coverage:
lidar dominates at close range, while radar covers most of the surveillance volume and serves as the primary sensor.

The dataset consists of two representative sequences.

\begin{figure}[t]
    \centering
    \includegraphics[width=\linewidth]{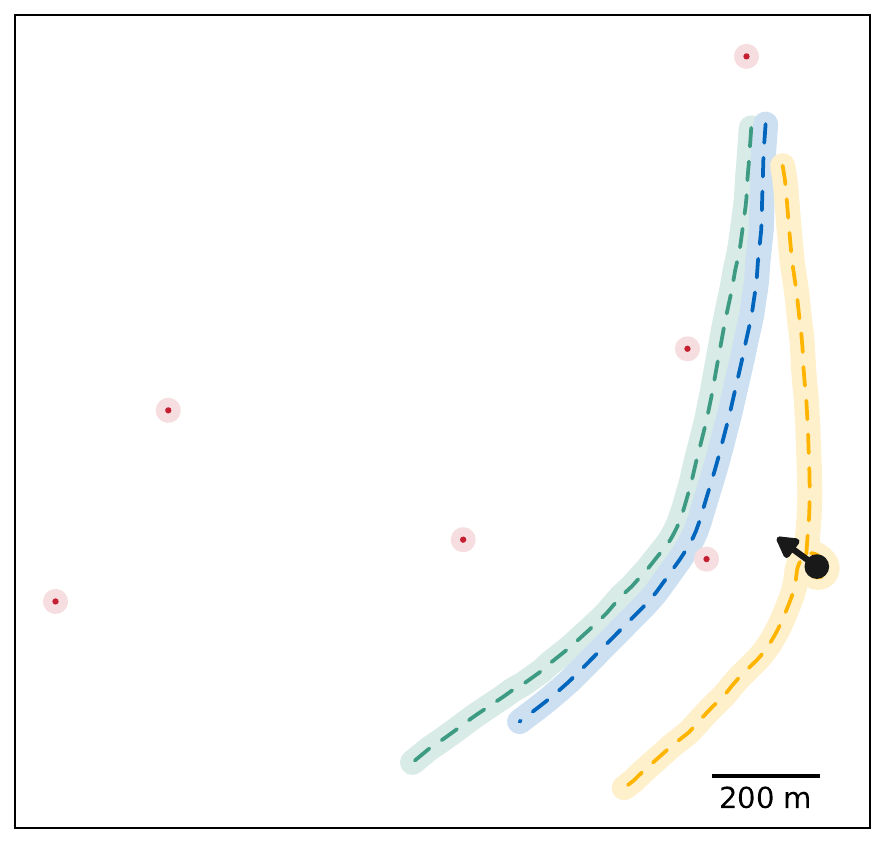}
    \caption{Scenario 1 ground-truth trajectories. Red markers show stationary objects (sea markers and islets). The ego-vessel remains fairly stationary at the position given by the black marker. The yellow target performs a close-range loop around the ego-vessel.}
    \label{fig:scenario1-gt}
\end{figure}

\textbf{Scenario 1: Mixed-Range Traffic.}
Three moving vessels operate alongside static sea markers and islets, illustrated in \cref{fig:scenario1-gt}.
Two of the vessels move at medium range ($1.5$~km to $300$~m), while one approaches to $\approx30$~m before departing.
This scenario stresses sensor handover across range regimes.

\begin{figure}[t]
    \centering
    \includegraphics[width=\linewidth]{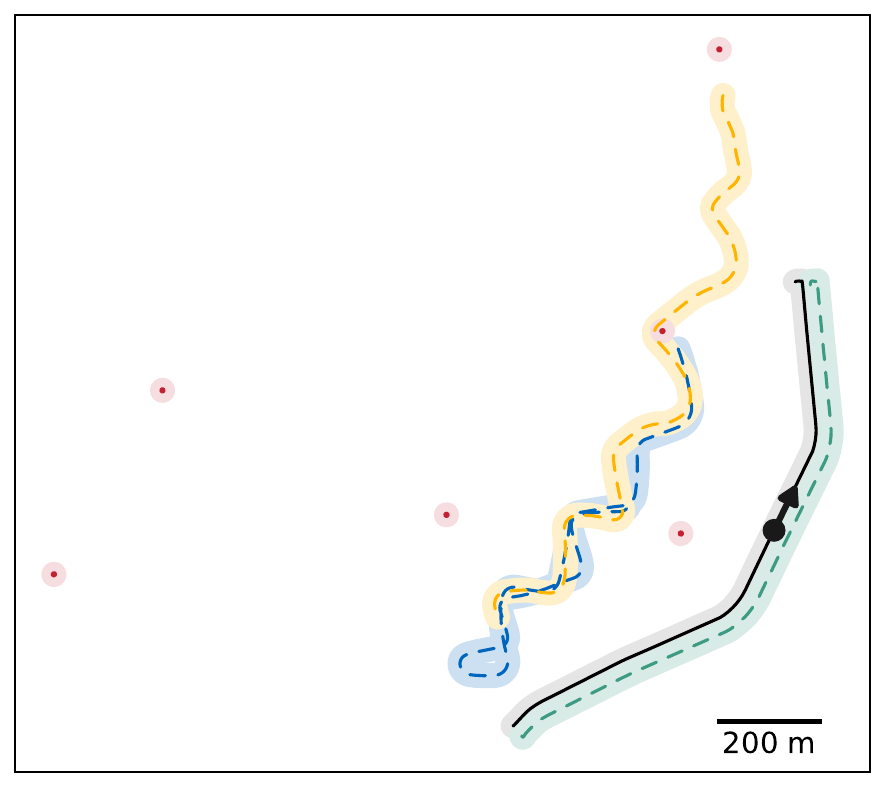}
    \caption{Scenario 2 ground-truth trajectories. Red markers show stationary objects (sea markers and islets). The ego-vessel follows the black trajectory in formation with a cooperating \ac{USV} (green trajectory).}
    \label{fig:scenario2-gt}
\end{figure}

\textbf{Scenario 2: Close-Range Formation with Occlusion.}
The ego vessel sails in close formation ($\approx30$~m separation) with a cooperating USV, while other vessels maneuver at medium range (\cref{fig:scenario2-gt}).
Targets intermittently disappear into the radar blind sector.
This scenario stresses association robustness under partial and time-varying observability.

\subsection{Tracker Configurations}
We evaluate the following configurations:
\begin{itemize}
\item \textbf{\acs{JPDA}:}
A standard \ac{JPDA} filter with heuristic $M$-of-$N$ track management and globally constant observability parameters.
\item \textbf{\acs{GM-PHD} (radar-only):}
A \ac{GM-PHD} filter using radar detections only.
\item \textbf{\acs{GM-PHD} (uniform):}
A \ac{GM-PHD} filter using globally constant detection probability and clutter intensity.
\item \textbf{\acs{GM-PHD} ($P_D$-aware):}
A \ac{GM-PHD} filter with state-dependent detection probability and constant clutter intensity.
\item \textbf{\acs{GM-PHD} (context-aware):}
A \ac{GM-PHD} filter with state-dependent detection probability and clutter intensity.
\end{itemize}

\begin{table}[t]
\centering
\caption{Tracker parameters used in all experiments.}
\label{tab:tracker-config}
\begin{tabular}{l l}
\toprule
\textbf{Component} & \textbf{Setting} \\
\midrule
Motion model & Constant velocity, $\sigma_{\mathrm{acc}}=0.8$ \\
State & $\vecx=[x,\dot{x},y,\dot{y}]^\top$ \\
Track init cov. & $\mathrm{diag}([10,15,10,15])^2$ \\
Meas./nav cov. & Varying; identical across trackers \\
\midrule
\multicolumn{2}{l}{\textbf{JPDA}} \\
Gating & $\chi^2$ gate, $P_G=0.95$ \\
$P_D$ & 0.4 \\
$\lambda$ & $10^{-3}$ \\
Initiation & min.\ 3 detections \\
Deletion & 30 missed updates \\
\midrule
\multicolumn{2}{l}{\textbf{GM-PHD (common parameters)}} \\
Gating & Mahalanobis dist. $< 3$ \\
Birth weight & $10^{-4}$ \\
Survival prob. & $1$ \\
Prune threshold & $10^{-6}$ \\
Merging & Mahalanobis dist. $< 4$ \\
Extraction threshold & $0.85$ \\
\midrule
\multicolumn{2}{l}{\textbf{GM-PHD (uniform and radar-only)}} \\
$P_D$ & 0.4 \\
$\lambda$ & $10^{-3}$ \\
\midrule
\multicolumn{2}{l}{\textbf{GM-PHD ($P_D$-aware)}} \\
$P_D$ & state-dependent, \cref{eq:pd-lidar,eq:pd-radar} \\
$\lambda$ & $10^{-3}$ \\
\midrule
\multicolumn{2}{l}{\textbf{GM-PHD (context-aware)}} \\
$P_D$ & state-dependent \cref{eq:pd-lidar,eq:pd-radar} \\
$\lambda$ & state-dependent \cref{eq:lambda-lidar,eq:lambda-radar} \\
\bottomrule
\end{tabular}
\end{table}

\Cref{tab:tracker-config} summarizes the tracker parameters used in all experiments.
All \ac{GM-PHD} variants share identical motion, birth, survival, pruning, merging, and extraction settings.
They differ only in the evaluation of detection probability $P_D$ and clutter intensity $\lambda$.
JPDA track confirmation and deletion were configured with enlarged $M$-of-$N$ windows to tolerate extended lidar non-detection sequences between radar updates.
Specifically, $N=30$ was required to preserve medium-range radar-supported tracks.

Based on the sensing geometry illustrated in \Cref{fig:sensor-coverage}, we now specify the state-dependent observability models used in the $P_D$-aware and context-aware configurations.

\paragraph{Lidar detection probability.}
The lidar provides reliable detection within short range but exhibits reduced performance near its effective boundary and cannot detect targets beyond its configured range.
We therefore define the lidar detection probability as a function of range $r$:
\begin{equation}
P_D^{\text{lidar}}(r) =
\begin{cases}
0.95, & 0 < r < 50 \\
0.2,  & 50 \le r < 80 \\
0,    & r \ge 80 .
\end{cases}
\label{eq:pd-lidar}
\end{equation}

\paragraph{Lidar clutter intensity.}
Small segmented regions are more likely to correspond to clutter than larger objects.
We therefore assign higher clutter intensity to small segmentation areas.
\begin{equation}
\lambda^{\text{lidar}} =
\begin{cases}
10^{-1}, & \text{area} < 10 \\
10^{-3}, & \text{otherwise} .
\end{cases}
\label{eq:lambda-lidar}
\end{equation}

\paragraph{Radar detection probability.}
Radar detection probability is constrained by range and the blind-sector shown in \cref{fig:sensor-coverage}.
Within the configured operational range, we assume constant detection probability, and zero otherwise:
\begin{equation}
P_D^{\text{radar}}(r,\theta) =
\begin{cases}
0.4, & 50 < r < 1612 \text{ and in-sector} \\
0,   & \text{otherwise} .
\end{cases}
\label{eq:pd-radar}
\end{equation}

\paragraph{Radar clutter intensity.}
Long-range radar detections are more prone to ambiguity and spurious returns.
We therefore increase clutter intensity beyond 1000 m:
\begin{equation}
\lambda^{\text{radar}}(r) =
\begin{cases}
10^{-3}, & r < 1000 \\
10^{-2}, & \text{otherwise} .
\end{cases}
\label{eq:lambda-radar}
\end{equation}

We process measurements at their native sensor update rates and handle each sensor scan as a batch at its associated timestamp.
No artificial temporal alignment or interpolation is applied.
Trackers therefore operate on naturally asynchronous measurement streams.

All trackers are constructed from unmodified Stone Soup components for prediction, update, and hypothesis management.
The underlying probabilistic update equations are unchanged across configurations.

\subsection{Evaluation Metrics}
Performance is assessed using \ac{HOTA}~\cite{luiten_hota_2021} and average \ac{GOSPA}~\cite{rahmathullah_generalized_2017}.

For \ac{HOTA}, Euclidean distance is used as the similarity measure.
Errors below 5 m are assigned full similarity, and similarity decreases linearly to zero at 30 m.
For \ac{GOSPA}, we use cutoff distance $c=30$ m and exponent $p=2$.
The cardinality weighting parameter is set to $\alpha=2$.

We evaluate all metrics at radar timestamps only.
For multi-sensor configurations, we ignore intermediate tracker outputs between radar updates to ensure fair comparison with the radar-only baseline.
This avoids inflating performance due to higher update frequency while preserving the temporal structure relevant to radar-supported tracking.
\section{Results}
This section quantifies the impact of global observability assumptions and evaluates the effect of context-aware modeling via \verb#DetectorContext#.
All metrics are averaged over both scenarios unless otherwise stated.

\subsection{Heuristic JPDA Provides Stable but Suboptimal Fusion}
\begin{figure*}[t]
    \centering
    \subfloat[JPDA]{%
        \centering
        \includegraphics[width=0.319\textwidth]{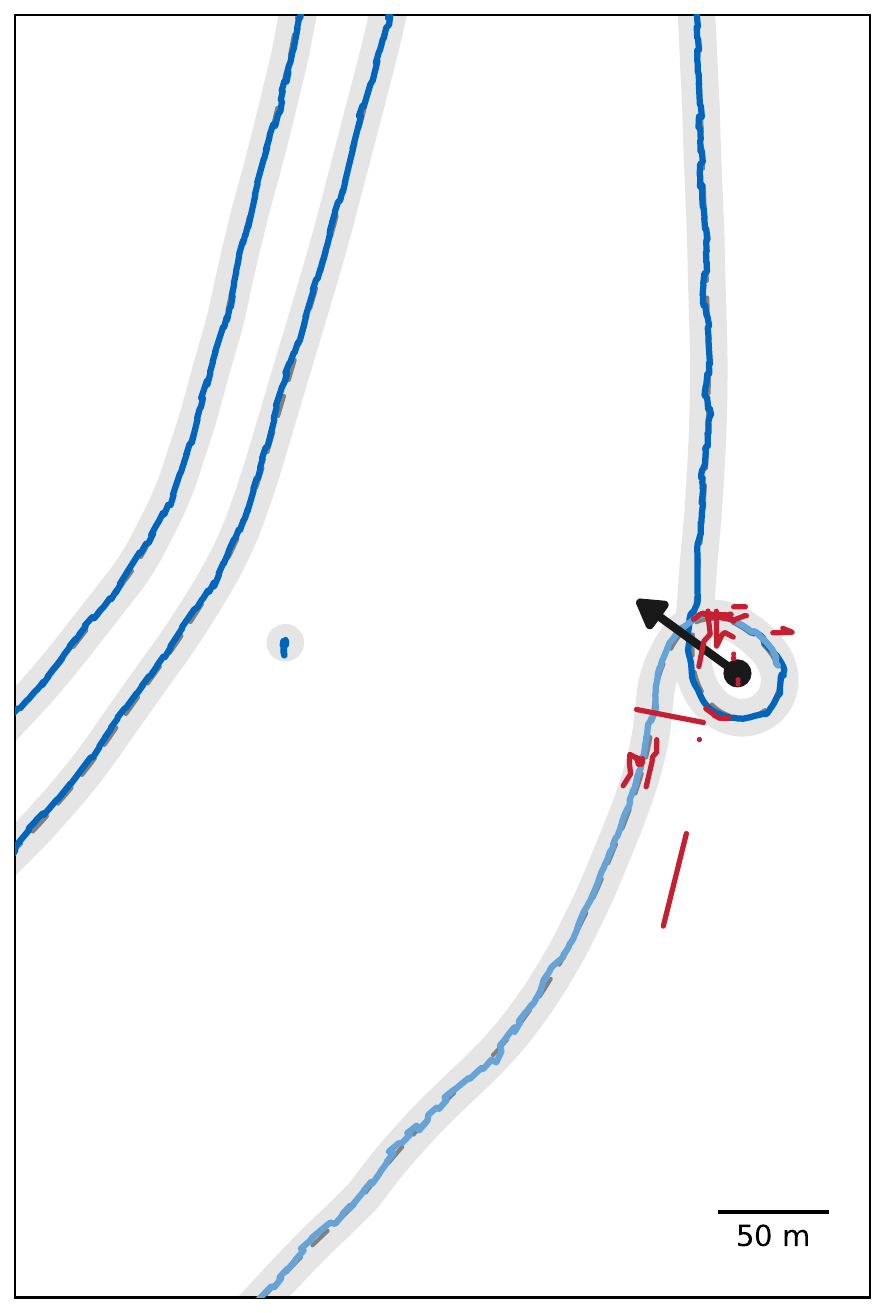}
        \label{fig:jpda-trajectories}
    }
    \hfill
    \subfloat[GM-PHD (uniform)]{%
        \centering
        \includegraphics[width=0.319\textwidth]{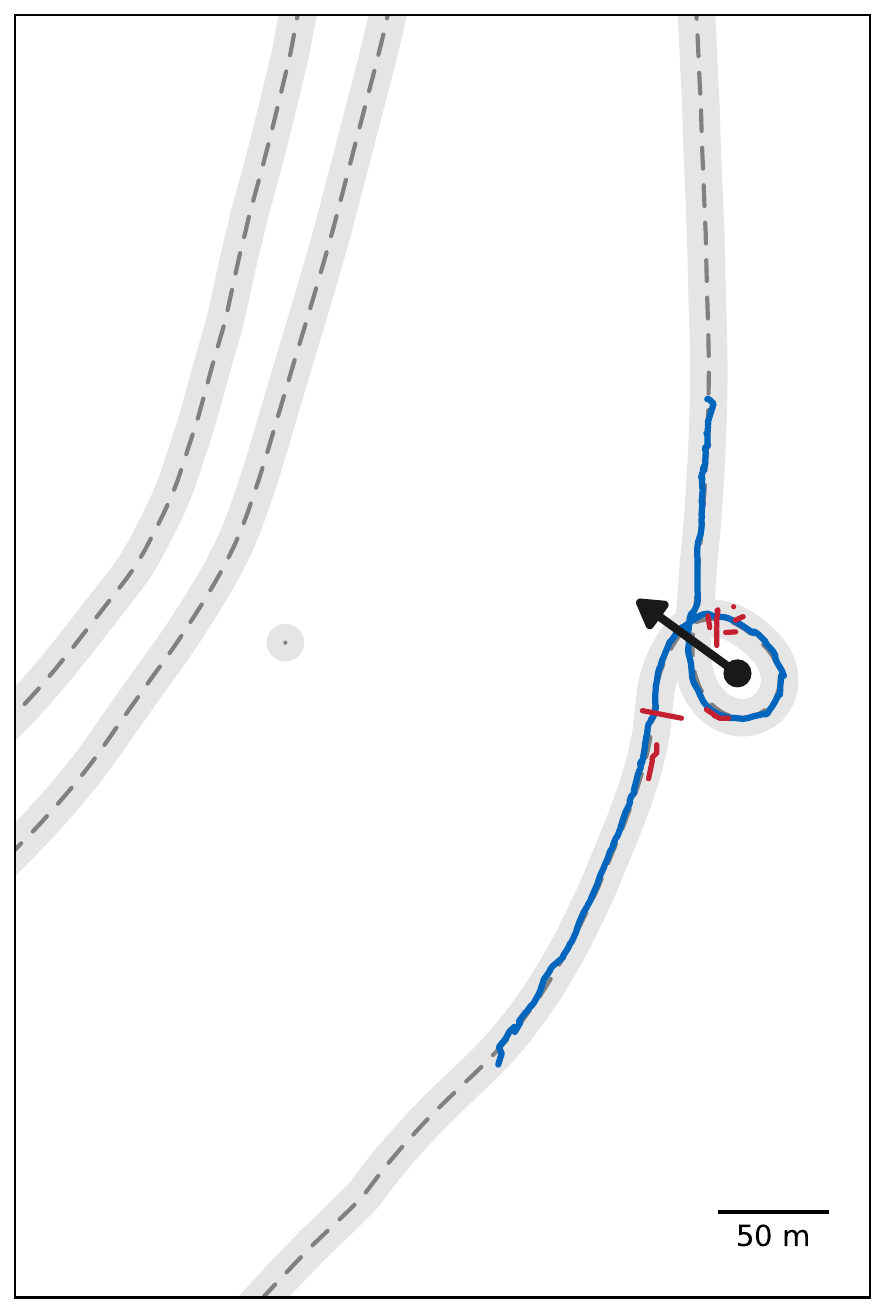}
        \label{fig:global-gmphd-trajectories}
    }
    \hfill
    \subfloat[GM-PHD (context-aware)]{%
        \centering
        \includegraphics[width=0.319\textwidth]{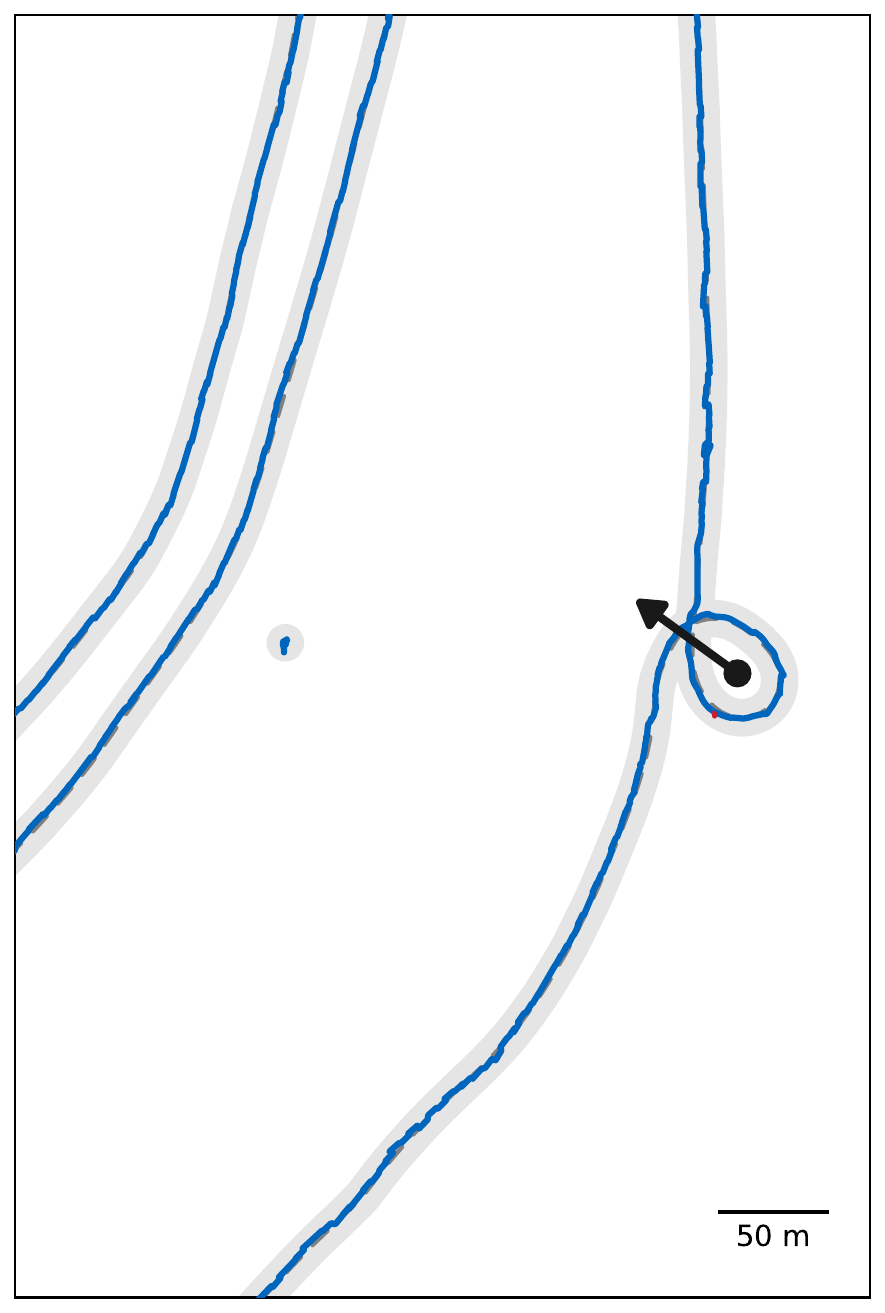}
        \label{fig:context-aware-gmphd-trajectories}
    }
    \caption{Representative tracking output for Scenario~1 over a 480~s interval.
Ground truth trajectories are shown as dashed gray lines.
Inlier target tracks are rendered in shades of blue, while clutter tracks are shown in red.
The ego-vessel position and heading are indicated by the black marker.
(a) JPDA maintains medium-range tracks but exhibits limited clutter suppression in lidar-dominated regions and fragments a target track within lidar range.
(b) GM-PHD with uniform observability loses radar-supported tracks due to accumulated missed detections.
(c) Context-aware GM-PHD restores track continuity across sensing regimes while effectively suppressing clutter.}
    \label{fig:tracker-trajectories}
\end{figure*}

\cref{fig:jpda-trajectories} shows representative tracking output on Scenario~1 for the baseline \ac{JPDA} configuration. 
Despite the need for an enlarged $M$-of-$N$ confirmation window, the filter maintains coherent medium-range tracks and successfully bridges radar-lidar handover.
Although occasional identity switches occur within sustained lidar tracking, at least one consistent track is maintained on each target for most of its lifetime.

\ac{JPDA} faces the tuning trade-off discussed in \cref{sec:intro}.
If confirmation logic is made too conservative, radar-supported tracks don't survive the high-rate lidar non-detections between radar updates.
If confirmation logic is relaxed to preserve those tracks, short-lived lidar clutter tracks are also allowed to persist.
No single confirmation setting simultaneously preserves medium-range radar tracks and aggressively suppresses close-range lidar clutter under globally uniform observability assumptions.

Quantitatively, \ac{JPDA} achieves a combined \ac{HOTA} score of 67.5 and an average \ac{GOSPA} of 35.2 (\cref{tab:results-summary}).
While medium-range tracking remains stable, overall performance is limited by lidar clutter persistence and occasional identity switches.

\begin{table*}[t]
\centering
\caption{Tracking performance across configurations. HOTA is computed using TrackEval multi-sequence aggregation. GOSPA is reported as RMS over all evaluation timesteps.}
\label{tab:results-summary}
\begin{tabular}{l SS SS SS}
\toprule
 & \multicolumn{2}{c}{Combined}
 & \multicolumn{2}{c}{Scenario 1} 
 & \multicolumn{2}{c}{Scenario 2} 
\\
\cmidrule(lr){2-3} \cmidrule(lr){4-5} \cmidrule(lr){6-7}
Method 
& \multicolumn{1}{c}{HOTA (\%) $\uparrow$}
& \multicolumn{1}{c}{GOSPA $\downarrow$}
& \multicolumn{1}{c}{HOTA (\%) $\uparrow$}
& \multicolumn{1}{c}{GOSPA $\downarrow$}
& \multicolumn{1}{c}{HOTA (\%) $\uparrow$}
& \multicolumn{1}{c}{GOSPA $\downarrow$} \\
\midrule
JPDA                             & 67.5 & 35.2 & 71.2 & 29.6 & 64.3 & 39.5 \\
GM-PHD (uniform)                 & 42.0 & 49.2 & 10.1 & 61.4 & 57.3 & 35.2 \\
GM-PHD (radar-only)              & 74.3 & 32.7 & 85.6 & 24.2 & 63.0 & 38.5 \\
\midrule
GM-PHD ($P_D$-aware)             & 80.1 & 29.4 & 89.5 & 22.1 & 71.4 & 34.6 \\
GM-PHD (context-aware) & \bfseries 81.4 & \bfseries 27.1 & \bfseries 89.7 & \bfseries 21.8 & \bfseries 73.5 & \bfseries 31.0 \\
\bottomrule
\end{tabular}
\end{table*}

\subsection{GM-PHD Fails Under Uniform Observability Assumptions}
The \ac{GM-PHD} filter relies exclusively on recursive Bayesian weight updates for track maintenance.
Under globally uniform detection probability, repeated high-rate lidar non-detections directly suppress the weights of radar-supported tracks between radar updates.

In Scenario~1, illustrated in \cref{fig:global-gmphd-trajectories}, the uniform \ac{GM-PHD} maintains tracks within the lidar-dominated region but fails to sustain radar-supported tracks.
This breakdown is reflected quantitatively in a \ac{HOTA} score of 10.1 for Scenario~1 and 42.0 overall, with an RMS \ac{GOSPA} of 49.2 (\cref{tab:results-summary}).
In contrast, the radar-only \ac{GM-PHD} achieves 85.6 HOTA on Scenario~1 and 74.3 HOTA combined.

Under uniform observability assumptions, multi-sensor fusion can become detrimental: the fused \ac{GM-PHD} configuration performs substantially worse than both the heuristic \ac{JPDA} baseline and the radar-only variant.

\subsection{Context-Aware Modeling Restores Fusion Performance}
With \verb#DetectorContext#, the systematic decay of radar-supported tracks under high-rate lidar non-detections disappears.
By conditioning detection probability on local sensor coverage, radar-supported tracks no longer accumulate spurious lidar non-detections. 
Track weights remain stable between radar updates, and continuity improves markedly during sensor handover across range regimes, as illustrated in \cref{fig:context-aware-gmphd-trajectories}.

As shown in \cref{tab:results-summary}, the context-aware \ac{GM-PHD} configuration substantially outperforms the uniform variant across all metrics.
It also surpasses both the heuristic \ac{JPDA} baseline and the radar-only configuration.
Most of the improvement arises from context-aware detection probability modeling, which prevents high-rate non-detections from suppressing valid tracks between radar updates.
Context-aware clutter modeling provides an additional improvement, increasing HOTA from 80.1 to 81.4.

\section{Discussion}
The results demonstrate that globally uniform observability assumptions induce structural failure under asynchronous and partially overlapping sensing. 
This failure does not arise from deficiencies in the underlying tracking algorithms, but from applying state-independent detection and clutter models in environments where observability varies spatially and temporally. 
When high-rate sensors repeatedly assert non-detections outside their effective coverage, probabilistic weight updates systematically suppress valid tracks.

Heuristic track management, as illustrated by \ac{JPDA}, can partially mask this effect by relaxing confirmation and deletion logic. 
However, such compensation introduces its own trade-offs, including reduced selectivity and prolonged clutter persistence. 
Heuristic track management can hide inconsistencies in the observability model.
Fully probabilistic trackers expose those inconsistencies.

The proposed \verb#DetectorContext# abstraction does not introduce new tracking theory, nor does it modify existing probabilistic update equations. 
Rather, it exposes locally conditioned observability information at the time of hypothesis formation while preserving modular separation between sensor modeling and track management. 
The abstraction relies on the tracker's existing formulation to exploit this locally conditioned context.

Extensions such as adaptive detection probability estimation or survival modeling are complementary but orthogonal to the architectural contribution presented here. 
\verb#DetectorContext# enables such methods but does not replace advanced probabilistic track management.

The context-aware detection and clutter models used in this study are intentionally simple and reflect known sensing geometry in the evaluated maritime dataset.
While the specific parameterization is scenario-dependent, the architectural abstraction itself is agnostic to the functional form of $P_D$ and $\lambda$.
Learned, adaptive, or data-driven observability models can be integrated through the same interface without modifying tracker equations.
The results therefore demonstrate not a dataset-specific tuning advantage, but the necessity of exposing observability as a first-class architectural concept.

Consequently, state-dependent observability modeling is not an optional feature; it is a fundamental architectural requirement for modern, multi-rate tracking frameworks.
\section{Conclusion}
Globally uniform detection probability and clutter intensity assumptions constitute a major source of structural failure in asynchronous, partially overlapping multi-sensor systems. 
When high-rate sensors assert non-detections outside their effective coverage, principled probabilistic filters suppress valid tracks, rendering fusion ineffective.

We introduced \verb#DetectorContext#, a framework-level abstraction that enables locally conditioned observability modeling in Stone Soup without re-deriving or modifying existing tracking algorithms. 
Experimental results on real radar-lidar data demonstrate that context-aware modeling restores stable fusion and allows probabilistic trackers to outperform heuristic alternatives.

Future work will extend \verb#DetectorContext# with richer environmental constraints, including occlusion reasoning, land masking, and radar shade. 
Because the abstraction decouples sensor modeling from tracking logic, learned observability parameters, target-dependent detection models, and platform pose uncertainty can be incorporated without modifying the underlying filters. 
Treating state-dependent observability as a first-class architectural component allows general-purpose frameworks to better align tracking theory with complex operational environments.

\section*{Acknowledgment}
This work was conducted in part within the NATO Science and Technology Organization (STO) Task Group SET-322, “Evaluation Framework for Multi-Sensor Tracking and Fusion Algorithms.”
The authors thank the group members for technical discussions that informed this research.

\bibliography{references}
\bibliographystyle{IEEEtran}

\end{document}